\documentclass[sigconf]{acmart}

\usepackage{amsmath}
\usepackage{url}
\usepackage{marvosym}
\usepackage{graphicx}
\usepackage{xcolor}
\usepackage{multirow}
\usepackage{tabularx}
\usepackage{algorithm}
\usepackage{algorithmic}
\usepackage{subcaption}
\usepackage{amsthm}
\usepackage{enumitem}

\newcolumntype{C}{>{\centering\arraybackslash}X} 

\usepackage{thmtools}
\usepackage{thm-restate}
\theoremstyle{plain}
\newtheorem{problem}{Problem}

\newcommand{\method}{TGPA}

\AtBeginDocument{%
  \providecommand\BibTeX{{%
    \normalfont B\kern-0.5em{\scshape i\kern-0.25em b}\kern-0.8em\TeX}}}

\copyrightyear{2025}
\acmYear{2025}
\setcopyright{acmlicensed}\acmConference[KDD '25]{Proceedings of the 31st ACM SIGKDD Conference on Knowledge Discovery and Data Mining V.2}{August 3--7, 2025}{Toronto, ON, Canada}
\acmBooktitle{Proceedings of the 31st ACM SIGKDD Conference on Knowledge Discovery and Data Mining V.2 (KDD '25), August 3--7, 2025, Toronto, ON, Canada}
\acmDOI{10.1145/3711896.3736845}
\acmISBN{979-8-4007-1454-2/2025/08}

\begin{document}

\title{Are You Using Reliable Graph Prompts? Trojan Prompt Attacks on Graph Neural Networks}
\author{Minhua Lin}
\authornote{Both authors contributed equally to this paper.}
\email{mfl5681@psu.edu}
\author{Zhiwei Zhang}
\authornotemark[1]
\email{zbz5349@psu.edu}
\affiliation{%
  \institution{The Pennsylvania State University}
  \city{University Park}
  \country{USA}
}

\author{Enyan Dai}
\email{enyandai@hkust-gz.edu.cn}
\affiliation{%
  \institution{Hong Kong University of Science and Technology (Guangzhou)}
  \city{Guangzhou}
  \country{China}}

\author{Zongyu Wu}
\email{zongyuwu@psu.edu}
\affiliation{%
  \institution{The Pennsylvania State University}
  \city{University Park}
  \country{USA}
}

\author{Yilong Wang}
\email{yvw5769@psu.edu}
\affiliation{%
  \institution{The Pennsylvania State University}
  \city{University Park}
  \country{USA}
}

\author{Xiang Zhang}
\email{xzz89@psu.edu}
\affiliation{%
  \institution{The Pennsylvania State University}
  \city{University Park}
  \country{USA}
}

\author{Suhang Wang}
\email{szw494@psu.edu}
\affiliation{%
  \institution{The Pennsylvania State University}
  \city{University Park}
  \country{USA}
}

\renewcommand{\shortauthors}{Minhua Lin et al.}

\ccsdesc[500]{Computing methodologies~Machine learning}



\keywords{Trojan Prompt Attacks; Graph Prompt Learning}


\received{20 February 2007}
\received[revised]{12 March 2009}
\received[accepted]{5 June 2009}


\begin{abstract}
Graph Prompt Learning (GPL) has been introduced as a promising approach that uses prompts to adapt pre-trained GNN models to specific downstream tasks without requiring fine-tuning of the entire model. Despite the advantages of GPL, little attention has been given to its vulnerability to backdoor attacks, where an adversary can manipulate the model's behavior by embedding hidden triggers. Existing graph backdoor attacks rely on modifying model parameters during training, but this approach is impractical in GPL as GNN encoder parameters are frozen after pre-training. Moreover, downstream users may fine-tune their own task models on clean datasets, further complicating the attack.
In this paper, we propose \method{}, a backdoor attack framework designed specifically for GPL. \method{} injects backdoors into graph prompts without modifying pre-trained GNN encoders and ensures high attack success rates and clean accuracy. To address the challenge of model fine-tuning by users, we introduce a finetuning-resistant poisoning approach that maintains the effectiveness of the backdoor even after downstream model adjustments. Extensive experiments on multiple datasets under various settings demonstrate the effectiveness of \method{} in compromising GPL models with fixed GNN encoders. Our code is publicly available at: {\url{https://github.com/ventr1c/TPGA}}.
\end{abstract}

\maketitle

\section{Introduction}
\label{sec:introduction}
Graph-structured data are prevalent in numerous real-world applications, such as social networks~\cite{hamilton2017inductive}, financial systems~\cite{wang2019semi}, and molecular graphs~\cite{wang2022molecular}. 
Graph Neural Networks (GNNs) have demonstrated great potential in modeling graphs through a message-passing mechanism~\cite{kipf2016semi,xu2018powerful}. Generally, traditional supervised GNNs heavily rely on abundant labeled data to perform well on downstream tasks. However, obtaining abundant high-quality labeled data is often expensive and labor-intensive. To address this issue, the ``pre-training and fine-tuning''~\cite{hu2020pretraining} paradigm is proposed, where GNNs are first pre-trained on unlabeled graphs using self-supervised learning (SSL) to capture intrinsic graph properties without task-specific labels. The pretrained GNNs are then adapted to test graphs for downstream tasks via fine-tuning. Despite the promising performance of such a paradigm, it suffers from several issues: (i) the significant gap between pre-training objectives, such as edge prediction~\cite{hu2020pretraining} in SSL tasks, and the specific requirements of downstream tasks like node or graph classification~\cite{You2020GraphCL,lin2023certifiably,xu2024llm}, will lead to suboptimal performance during fine-tuning~\cite{liu2023pre}; and (ii) they are prone to forget previously learned knowledge during fine-tuning, leading to severe catastrophic forgetting~\cite{liu2021overcoming}.

Recently, graph prompt learning (GPL) has emerged as a promising solution to address the aforementioned issues. To avoid catastrophic forgetting, the pretrained GNN is frozen. Instead, graph prompts and task headers are learned to align the pre-trained GNN encoder with specific downstream tasks. Generally, prompts are learnable tokens~\cite{liu2023graphprompt,Fang2023UniversalPT} or graphs~\cite{sun2023graph} for better adaptation and task headers are linear layers with few parameters for the downstream tasks. This approach enables more effective and efficient adaptation of pre-trained GNNs to new tasks by leveraging prompts to bridge the gap between pre-training and downstream task objectives, showing promising results. For example, GraphPrompt~\cite{liu2023graphprompt} and All-in-One~\cite{sun2023all} introduce a learnable prompt vector in the latent space and a learnable subgraph for downstream tasks, respectively, which are then fed into the pre-trained model, allowing better adaptation to the downstream task and data. 




Despite their promising results, like GNNs~\cite{zhang2021backdoor,xi2021graph,dai2023unnoticeable,zhang2023graph,lyu2024crossba,zhang2025robustness}, GPL methods might be vulnerable to graph backdoor attacks. Generally, graph backdoor attack attaches backdoor triggers to a small set of nodes in a training graph and labels those nodes as the target class. These triggers can be predefined subgraphs or synthesized by a trigger generator. A GNN classifier trained on the poisoned graph will misclassify test nodes attached with triggers to the target class while behaving normally on clean nodes. Backdoor attacks have posed significant threats to GNN classifier learning, putting the adoption of graph prompt at a similar risk. In particular, the adopted graph prompts are often shared or publicly released by others, which could be a backdoored prompt. This leads to significant risks in high-stakes applications such as fraud detection within transaction networks. Specifically, the adversary could inject backdoor triggers into the network, allowing fraudulent transactions to evade detection with the attacked prompts. Thus, graph backdoor attacks are an important and increasingly concerned problem.

However, existing graph backdoor attacks mainly focus on GNNs. The backdoor vulnerability of GPL is underexplored. Adapting existing graph backdoor attacks to GPL presents significant challenges: \textbf{(i)} existing graph backdoor attacks typically rely on modifying model parameters by training GNN encoders on poisoned datasets by backdoor triggers. However, in GPL, the GNN encoder is frozen after pre-training, and attackers do not have access to the pre-training process. This is because the pretrained GNN encoders are usually maintained and released by trustworthy third parties. Modifying the encoder would not only require significant resources but could also be easily observed by these third parties, rendering such attacks impractical. The only components available to the attacker are the graph prompts and a small downstream task header with limited or no learnable parameters, making it difficult to directly apply existing attacks to GPL; \textbf{(ii)} A straightforward approach would be to extend existing backdoor attacks to GPL by updating the downstream task headers on poisoned datasets. However, our preliminary results in Sec.~\ref{sec:limitation_existing_GBA} show that
this approach is ineffective. The task header usually has limited parameters~\cite{sun2023all,Fang2023UniversalPT} or even parameters-free~\cite{liu2023graphprompt}, which tend to overfit either to clean features or trigger features \cite{gu2023gradient, liu2024does}, leading to a poor balance between attack performance and model utility. Moreover, as downstream users may train their own task headers, it becomes even more difficult for attackers to inject backdoors into task headers.

Therefore, in this paper, we study a novel problem of graph backdoor attack on GPL where model parameters of GNN encoders are frozen. There are two major challenges: (\textbf{i}) With the GNN encoder frozen, how to achieve the backdoor attack goal while maintaining model utility on clean samples? (\textbf{ii}) Downstream users may fine-tune their task headers or graph prompts on clean datasets to improve performance, significantly degrading backdoor attack success rate. How to ensure the attack’s effectiveness even if the trojan graph prompts and downstream task models are fine-tuned on clean datasets by end users?
In an effort to address the aforementioned challenges, we propose a novel framework named \textit{\underline{T}rojan \underline{G}raph \underline{P}rompt \underline{A}ttack}  (\method{}). To overcome the challenge of ensuring both backdoor attack performance and model utility without poisoning the pretrained GNN encoders, \method{} introduces trojan graph prompt attacks to make the task header learn the association between \textit{(trojan graph prompt, trigger)} with the \textit{target class}. Hence, the task header will misclassify nodes attached with both trojan graph prompt and trigger to the target class but behavior normally when only trojan graph prompt is attached to a node. Trojan graph prompts are trained from clean graph prompts on poisoned datasets that include trigger-attached graphs. A feature-aware trigger generator is employed in \method{} to generate powerful triggers that achieve both high attack success rates and strong clean accuracy. To maintain the effectiveness of trojan graph prompt attacks after fine-tuning, we introduce a fine-tuning-resistant graph prompt poisoning approach that ensures trojan graph prompts remain functional even after fine-tuning. 

In summary, our \textbf{main contributions} are: (i) We study a novel backdoor attack problem targeting GPL without poisoning GNN encoders; (ii) We propose \method{}, a novel trojan graph prompt attack framework that injects backdoors into graph prompts, achieving both high attack success rates and clean accuracy, and resisting fine-tuning of graph prompts and downstream task models; and (iii) Extensive experiments on various datasets under different settings demonstrate the effectiveness of \method{} in backdooring GPL.

\section{Related Work}
\textbf{Graph Prompt Learning}. 
Recently, graph prompt learning (GPL), which adopts ``pretrain-prompt-finetuning'' strategy, has emerged as a promising alternative to traditional ``pretrain-finetuning'' approaches~\cite{sun2023all, zhao2024all, sun2023graph, li2024zerog, tan2023virtual, yu2024generalized, yu2024hgprompt,li2025fairness}. Generally, GPL pretrain a GNN encoder on unlabeled graphs. To adapt the pretrained GNN encoder to a target graph for downstream tasks, GPL introduces graph prompts to better align the input and task for GNN encoder and add a task header on top of the encoder for downstream tasks. The graph prompts and task header are learned to enhance alignment with pre-trained GNN models while the pretrained GNN encoder is frozen to avoid cartographic forgetting. 
Generally, GPL methods can be categorized into two categories: (i) \textit{prompt-as-tokens}~\cite{fang2022prompt, fang2024universal, chen2023ultra, gong2023prompt, liu2023graphprompt, ma2024hetgpt, shirkavand2023deep, sun2022gppt}, which treats graph prompts as additional features appended to the graph. For example, GraphPrompt~\cite{liu2023graphprompt} introduces a learnable prompt vector in the latent space for downstream tasks. 
(ii) \textit{prompt-as-graphs}~\cite{ge2023enhancing, huang2024prodigy, sun2023all}, which introduces prompt tokens as additional nodes with the same representation size as the original nodes~\cite{ge2023enhancing, huang2024prodigy, sun2023all}. For instance, All-in-One~\cite{sun2023all} creates a learnable subgraph of prompt tokens, where these tokens have both inner and cross-links to the original graph nodes. This augmented graph is then fed into the pre-trained model, allowing better integration with the task-specific data.
More details are in Appendix~\ref{appendix:more_related_works_GPL}.

\noindent\textbf{Backdoor Attacks on Graph}. 
Generally, graph backdoor attacks~\cite{zhang2021backdoor, xi2021graph, dai2023unnoticeable, zhang2024dpgba, zhang2023graph,lyu2024crossba,song2024krait,hou2024adversarial} inject backdoor triggers to training graphs and assign target class to nodes attached with triggers. A GNN trained on the poisoned graph will then be misled to predict test nodes/graphs attached with the trigger to the target class. 
In early efforts, 
SBA~\cite{zhang2021backdoor} injects universal triggers into training data through a subgraph-based method. 
Later, UGBA~\cite{dai2023unnoticeable} adopts an adaptive trigger generator to produce unnoticeable backdoor triggers with high cosine similarity to the target node. CrossBA~\cite{lyu2024crossba} is a cross-context backdoor attack against GPL, which optimizes the trigger graph and poisons the pretrained GNN encoder.
Despite their strong attack performance, these poisoning-based backdoor attacks require training the backdoored model on a poisoned dataset with trigger-attached samples, which is impractical in trojan graph prompt attacks without poisoning the GNN encoder. Krait~\cite{song2024krait} is a concurrent work against GPL without poisoning GNN encoders. However, it mainly targets All-in-One~\cite{sun2023all}, which uses subgraphs as prompts, making it ineffective for methods employing learnable tokens as prompts. Moreover, it also fails to address scenarios where graph prompts and task headers are fine-tuned, which can significantly impact attack effectiveness.
Our proposed method is inherently different from these methods as \textbf{(i)} we propose a trojan graph prompt attack to conduct backdoor attacks against various GPLs without poisoning GNN encoders; \textbf{(ii)} we design a novel finetuning-resistant graph prompt poisoning such that our trojan graph prompt attack remains effective after fine-tuning the graph prompts and task headers. More details are in Appendix~\ref{appendix:more_related_works_backdoor}.




\section{Backgrounds and Preliminaries}
\subsection{Notations}
\label{sec:notations}
Let $\mathcal{G}=(\mathcal{V},\mathcal{E}, \mathbf{X})$ denote a graph, where $\mathcal{V}=\{v_1,\dots,v_N\}$ is the set of $N$ nodes, $\mathcal{E} \subseteq \mathcal{V} \times \mathcal{V}$ is the set of edges, and $\mathbf{X}=\{\mathbf{x}_1,...,\mathbf{x}_N\}$ is the set of node attributes with $\mathbf{x}_i\in\mathbb{R}^{1\times{d}}$ being the node attribute of $v_i$. $d$ is the size of node attributes. $\mathbf{A} \in \mathbb{R}^{N \times N}$ is the adjacency matrix of $\mathcal{G}$, where $\mathbf{A}_{ij}=1$ if nodes ${v}_i$ and ${v}_j$ are connected; otherwise $\mathbf{A}_{ij}=0$. In this paper, we focus on graph trojan attacks against GPL, targeting node-level downstream tasks. 
We consider the few-shot learning setting for downstream tasks. In this setting, $\mathcal{G}$ includes only a limited subset of labeled nodes $\mathcal{V}_L \subseteq \mathcal{V}$, where $\mathcal{V}_L = \{y_1, \dots, y_{N_L}\}$ are the labels of $N_L$ nodes, and each class is provided with $k$ labeled examples. The remaining nodes are unlabeled, denoted as $\mathcal{V}_U$. The test nodes are a subset of the unlabeled nodes, denoted as $\mathcal{V}_T \subset \mathcal{V}_U$, with $\mathcal{V}_T \cap \mathcal{V}_L = \emptyset$. 

\subsection{Threat Model}
\subsubsection{Attacker's Goal}
The goal of the adversary is to mislead the task header built upon the GNN encoder to classify target nodes attached with backdoor triggers as target class. Simultaneously, the task header should behave normally for clean nodes without triggers attached. Specifically, in the scenario of ``pretraining, prompting and fine-tuning'', the GNN encoder is pretrained in a clean dataset and its model parameters are then frozen. The adversary aims to maliciously craft trojan prompts and backdoor triggers for the graph such that the attacked task header will build an association between the trigger-attached nodes with the trojan prompt applied and the target class in the downstream task.

\begin{table}[t]
    \centering
    \caption{Results of the extensions of existing graph backdoor attacks in GPL(Attack Success Rate (\%)|Clean Accuracy (\%))}
    \small
    \vskip -1em 
    \begin{tabularx}{0.98\linewidth}{p{0.1\linewidth}CCCC}
    \toprule 
    Datasets & Clean Graph & SBA-P & GTA-P & UGBA-P \\
    \midrule
    
    Cora & 73.7 & 70.4|23.8 & 72.3|28.8 & 72.2|28.9 \\
    Pubmed &  71.0 & 63.8|31.4 &  63.8|21.2 & 70.4|17.3 \\
    \bottomrule
    \end{tabularx}
    \vskip -1em
    \label{tab:preliminary_result_GBA}
\end{table}

\subsubsection{Attacker's Knowledge and Capability} 
To conduct backdoor attacks against GPL without poisoning GNN model parameters, we assume that the attacker has no access to the pretraining phase but full access to prompting and fine-tuning phases. Specifically, the attacker can access the pretrained GNN encoder, but the training set and the pretraining method of the GNN encoder are unknown to the attacker. In the prompting and fine-tuning phases, all the information, including GPL strategies, the downstream tasks, and the corresponding datasets, is available to the attacker. The attacker is capable of crafting a trojan graph prompt and attaching triggers and labels to nodes to poison graphs during the prompting and fine-tuning phases. During the inference phase, the attacker can apply the trojan prompts and attach triggers to the target test node. 

We argue this is a reasonable and practical setting as model providers usually pretrain the model privately and release the pretrained model to customers for downstream applications. Moreover, unlike prompts in NLP field, which are usually in discrete textual format~\cite{taylor2020autoprompt,gao2021making}, graph prompts are usually learnable tokens~\cite{liu2023graphprompt,sun2022gppt,Fang2023UniversalPT,yu2024multigprompt} or augmented graphs~\cite{sun2023all,huang2024prodigy}, which are challenging for end users to understand and interpret them. Therefore, the attacker may act as a malicious third-party service provider to provide the trojan graph prompts and backdoored task header for the users to achieve the attack goal in various downstream tasks.

\subsection{Limitations of Existing Graph Backdoor Attacks against GPL}
\label{sec:limitation_existing_GBA}
To validate the simple adpation of existing graph backdoor attacks against GPL is ineffective,
we extend three representative graph backdoor attacks, i.e., SBA~\cite{zhang2021backdoor}, GTA~\cite{xi2021graph} and UGBA~\cite{dai2023unnoticeable}, {to attack GPL with the GNN encoder frozen}. Detailed descriptions of these methods are in Sec.~\ref{sec:compared_methods}. The architecture of the pretrained GNN encoder is GCN~\cite{kipf2016semi}. The pretraining method is SimGRACE~\cite{xia2022simgrace}. The GPL method is set as GraphPrompt~\cite{liu2023graphprompt}. 
Other settings are the same as the evaluation protocol in Sec.~\ref{sec:evaluation_protocol}. Experimental results on Cora and Pubmed~\cite{sen2008collective} datasets are presented in Table~\ref{tab:preliminary_result_GBA}. From the table, we observe that all the compared methods achieve only up to $32\%$ ASR, which validates the ineffectiveness of simply extending existing attack methods.

\subsection{Problem Statement}
Our preliminary analysis in Sec.~\ref{sec:limitation_existing_GBA} verifies that simply adapting existing graph backdoor attacks in GPL is ineffective. Thus, we propose to study a novel trojan graph prompt attack against GPL without poisoning GNN encoders. Specifically, we aim to carefully craft trojan graph prompts and triggers such that the task header will associate (trojan graph prompt, trigger) with the target class
The trojan graph prompt attack is formally defined as:
\begin{problem}[Trojan Graph Prompt Attack]
    Given a pretrained GNN encoder $h_{\theta}$ and a clean attributed graph $\mathcal{G}=(\mathcal{V},\mathcal{E},\mathcal{X})$ on the downstream task $\tau$, where $\mathcal{V}_L$ is a set of labeled nodes with corresponding labels $\mathcal{Y}_L$, $\mathcal{V}_P$ is the set of nodes to attach triggers and labels, we aim to learn a trojan graph prompt $p^{*}$, a task header $f_\tau$ built upon $h_{\theta}$ for $\tau$ and a trigger generator $f_g(v_i)\rightarrow g_i$ to produce triggers to ensure $f_{\tau}$ classify the test node attached with the trigger to target class $y_t$, even after $f_{\tau}$ and $p$ are fine-tuned. The objective function can be written as: 
    \begin{equation}
    \label{eq:problem_objective}
    \begin{aligned}
        \min_{\theta_g} &\sum_{v_i\in\mathcal{V}_U\backslash\mathcal{V}_T}l\left(f_{\tau}\{h_{\theta}[a_{g}(v_i^{p^*},g_i)]\}, y_t\right) \\
        s.t.\ p^{*},\theta^*_{\tau}=&\underset{p}{\arg\min}\sum_{v_i\in\mathcal{V}_L}l\left(f_{\tau}\{h_{\theta}[v_i^p]\},y_i\right) \\
        & + \sum_{v_i\in\mathcal{V}_P}l\left(f_{\tau}\{h_{\theta}[a_g(v_i^p, g_i)]\},y_t\right) 
    \end{aligned}
    \end{equation}
    where $l(\cdot)$ represents the cross entropy loss, and $\theta_g$ denotes the parameters of trigger generator $f_g$. 
    $a_g$ and $a_p$ are the trigger attachment and graph prompting operations, respectively, with $v_i^p = a_p(p,v_i)$. Specifically, $a_p(p,v_i)$ denotes the attachment of the graph prompt $p$ to $v_i$, $a_{g}(v_i^{p},g_i)$ denotes the process where the graph prompt-embedded node $v_i^{p}$ is further attached to trigger $g_i$, which is generated by $f_g(v_i)$.
\end{problem}
Unlike~\cite{dai2023unnoticeable,zhang2024dpgba}, we do not assume that $|\mathcal{V}_P|$ must be constrained by a specific attack budget. This is because the attacker will only release the trojan graph prompts and the task header to the downstream users and keep the poisoned graph private. For instance, consider a real-world scenario, where the users download trojan graph prompts and an attacked task header from an untrusted third party for downstream task $\tau_j$ on graph $\mathcal{G}$. The trojan graph prompt and the attacked task header are usually trained on a poisoned graph $\mathcal{G}'$ privately owned by the adversary. Therefore, the adversary can select any number of poisoned samples for enhanced attack performance.
During the inference phase, once the users use the trojan graph prompt and attacked header, the adversary can then inject triggers to the clean graph $\mathcal{G}$ to conduct backdoor attacks.

\section{Methodology}
\begin{figure}
    \centering
    \includegraphics[width=0.98\linewidth]{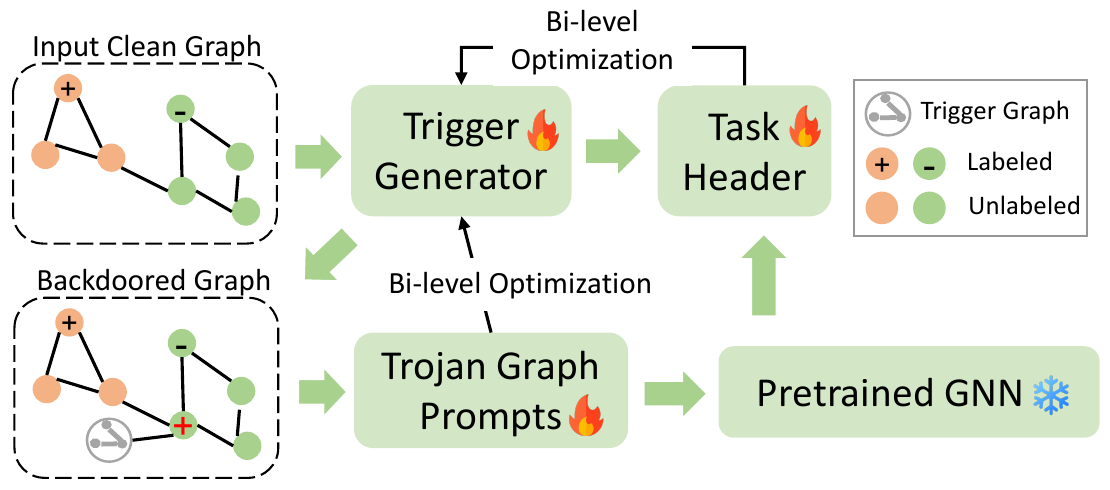}
    \vskip -1em
    \caption{An overview of proposed \method{}.}
    \vskip -1em
    \label{fig:framework_TGPA}
\end{figure}

In this section, we present the details of our \method, which aims to conduct backdoor attacks against GPL without poisoning GNN model parameters. There are mainly two challenges to be addressed: \textbf{(i)} How to design the trojan graph prompt and trigger to increase backdoor attack performance on trigger-attached samples without affecting the clean accuracy of benign samples.  
\textbf{(ii)} How to ensure the attack performance even the task headers are fine-tuned in a clean dataset by users?
To address these challenges, a novel framework of {\method} is proposed, which is illustrated in Fig.~\ref{fig:framework_TGPA}. {\method} is composed of a clean prompt generator, a trojan prompt generator, and a trigger generator.  Specifically, to address the first challenge, a clean graph prompt and a benign task header are first trained to maintain high accuracy for benign samples. A feature-aware trigger optimization and a task-specific trojan prompt poisoning are then proposed to train a trigger generator and inject backdoor to the prompt to ensure high clean accuracy and attack success rate simultaneously. To address the second challenge, we propose a novel finetuning-resistant graph prompt poisoning that ensures the effectiveness of the backdoor in trojan graph prompt even when the task header undergoes fine-tuning.

\subsection{Backdooring Graph Prompts}

\subsubsection{Training Clean Prompt and Task Header}
\label{sec:train_clean_prompt}
Due to the conflicting goals of achieving high attack success rate (ASR) and clean accuracy (CA) in the objective of Eq.~(\ref{eq:problem_objective}), simultaneously optimizing trojan graph prompts and triggers are prone to boost ASR at the cost of reduced CA or vice versa, leading to instability and challenges in achieving high ASR and CA at the same time. To alleviate this issue, we propose to separate the training of clean graph prompts and the generation of triggers, with the clean graph prompts generation conducted first, followed by trigger generation. Specifically, to ensure the task header $f_{\tau}$ built upon $h_{\theta}$ can maintain high accuracy on benign samples, in the first step, we train a clean graph prompt $p$ and $f_{\tau}$ as
\begin{equation}
\begin{aligned}
\label{eq:train_clean_prompt_header}
    & \min_{p,\theta_\tau}\mathcal{L}_c=\sum_{v_i\in\mathcal{V}_L}l\left(f_\tau[h_\theta\{a_p(p, v_i)\}], y_i\right), 
\end{aligned}
\end{equation}
where $l(\cdot)$ is the cross entropy loss, $\theta_\tau$ is the parameters of the task header $f_\tau$, $a_p$ is the graph prompting operation. $\mathcal{V}_L=\{v_1,\ldots,v_{N_L}\}$ is the set of labeled nodes in the downstream task. By solving Eq.~(\ref{eq:train_clean_prompt_header}), a good initialization of graph prompt $p$ and task header $f_{\tau}$ is obtained, respectively, to ensure high accuracy
in the clean dataset. 

\subsubsection{Feature-aware Trigger Generation}
\label{sec:trigger_generation}
Once the clean prompt graph $p$ and task header $f_{\tau}$ are trained, the next step is to generate triggers to poison the dataset. To make the generated triggers more effective and flexible, inspired by~\cite{dai2023unnoticeable}, instead of using predefined triggers, we propose an adaptive trigger generator to generate triggers based on the representations of the target node after prompting. Specifically, given a pretrained GNN encoder $h_{\theta}$, a node $v_i$ and a clean graph prompt $p$, we apply an MLP as the adaptive trigger generator $f_g$ to simultaneously generate node features and structure of the trigger for node $v_i$ by:
\begin{equation}
\begin{aligned}
    \label{eq:trigger_generator}
    \mathbf{h}_i^m = \text{MLP}\left(h_{\theta}\{a_p(p,v_i)\}\right),\ \ \ \mathbf{X}_i^g = \mathbf{W}_f\cdot\mathbf{h}_i^m,\ \ \ \mathbf{A}_i^g = \mathbf{W}_a\cdot\mathbf{h}_i^m,
\end{aligned}
\end{equation}
where $\mathbf{W}_f$ and $\mathbf{W}_a$ are learnable parameters for feature and structure generation, respectively. $\mathbf{X}_i^g\in\mathbb{R}^{s\times{d}}$ is the feature matrix of trigger nodes, where $s$ and $d$ represent the size of the generated trigger and the dimension of trigger features, respectively. $\mathbf{A}_i^g\in\mathbf{R}^{s\times{s}}$ is the adjacency matrix of the generated trigger. As real-world graphs are generally discrete, we binarize the $\mathbf{A}_i^g$ in the forward computation to align with the binary structure of the graph; while the continuous value is used in the backward propagation. 


\subsubsection{Trojan Graph Prompt Poisoning}
\label{sec:trojan_graph_prompt_poisoning}
A key to successful graph backdoor attacks is the capability of the backdoored model in correlating attached triggers with the target class. Traditional graph backdoor attacks~\cite{zhang2021backdoor,xi2021graph,dai2023unnoticeable,lyu2024crossba} generally rely on training GNNs in the poisoned dataset to build such a correlation. However, in our setting, the GNN encoder is pretrained in a clean dataset and its model parameters are frozen, making it difficult to build the connection. Our preliminary analysis in Sec.~\ref{sec:limitation_existing_GBA} also verifies the ineffectiveness of simply extending these attacks by poisoning the task headers. To enable backdoor attack against GPL with fixed GNN encoders, we propose a novel attack method to craft a trojan graph prompt and an attacked task header by further training the clean graph prompt $p$ and task header $f_{\tau}$ in the poisoned dataset. Specifically, we first randomly select a small set of nodes from $\mathcal{V}_U$ to attach triggers and denote the set as $\mathcal{V}_P$. With the clean graph prompt $p$ and the adaptive trigger generator $f_g$ in Sec.~\ref{sec:trigger_generation},  we train the graph prompt and the task header $f_\tau$ in the poisoned dataset as:
\begin{equation}
\label{eq:train_prompt_header_in_poisoned_dataset}
\begin{aligned}
    \min_{p,\theta_{\tau}}\mathcal{L}_{p}(\theta_g,p,\theta_{\tau}) & = \sum_{v_i\in\mathcal{V}_L}l\left(f_{\tau}[h_{\theta}\{a_p(p,v_i)\}],y_i\right)\\
   &+ \sum_{v_i\in\mathcal{V}_P}l\left(f_{\tau}[h_{\theta}\{a_g(a_p(p,v_i), g_i)\})],y_t\right),\\
\end{aligned}
\end{equation}
where $a_g$ is the trigger attachment operation and $g_i=(\mathbf{X}_i^g,\mathbf{A}_i^g)$ is the trigger generated by $f_g$ based on Eq.~(\ref{eq:trigger_generator}). As we already have good initialization on both clean graph prompt $p$ and task header $f_{\tau}$ based on Eq.~(\ref{eq:train_clean_prompt_header}), Eq.~(\ref{eq:train_prompt_header_in_poisoned_dataset}) ensures the attack performance while maintaining high accuracy.

The trigger generator $f_g$ will then be optimized to effectively mislead the graph prompt $p$ to make the fixed GNN encoder $h_{\theta}$ to produce embeddings for various nodes from $\mathcal{V}_U$ that are predicted to $y_t$ by the task header $f_{\tau}$ once attached with generated triggers and prompted with $p$. Formally, the objective is given as:
\begin{equation}
    \label{eq:trigger_optimization}
    \min_{\theta_g}\mathcal{L}_{g}(\theta_g,p,\theta_{\tau}) = \sum_{v_i\in\mathcal{V}_U}l\left(f_{\tau}[h_{\theta}\{a_g(a_p(p,v_i), g_i)\})],y_t\right).
\end{equation}


Combine Eq.~(\ref{eq:train_prompt_header_in_poisoned_dataset}) and Eq.~(\ref{eq:trigger_optimization}) together, we apply a bi-level optimization to optimize the trigger generator, trojan graph prompt and task header together such that the adaptive trigger generator can successfully attack the pretrained GNN encoder with the trojan graph prompt, which can be formulated as:
\begin{equation}
\label{eq:bi_level_optimization_wo_finetuning_resistant}
    \min_{\theta_g}\mathcal{L}_g(\theta_g,p^*,\theta_{\tau}^*)  \quad s.t. \quad p^*,\theta_{\tau}^*=\underset{p,\theta_{\tau}}{\arg\min}~{\mathcal{L}_p(\theta_g,p,\theta_{\tau})}.
\end{equation}
By solving Eq.~(\ref{eq:bi_level_optimization_wo_finetuning_resistant}), a backdoor is then injected to the trojan graph prompt to build the association between trigger and target class.


\subsection{Finetuning-Resistant Graph Prompt Poisoning}
\begin{table}[t]
    \centering
    \caption{Attack results of on Cora before and after fine-tuning task header (ASR (\%)|CA (\%))}
    \small
    \vskip -1em 
    \begin{tabularx}{1\linewidth}{lcCCCCc}
    \toprule 
    Finetune & Clean Graph & SBA-P & GTA-P & UGBA-P & \method{}/R \\
    \midrule
    
    Before & 73.7 & 70.4|23.8 & 72.3|28.8 & 72.2|28.9 & 73.1|88.6 \\
    After &  67.6 & 65.6|13.7 & 65.7|16.4  & 65.6|3.4 & 66.5|41.0\\
    \bottomrule
    \end{tabularx}
    \vskip -1em
    \label{tab:preliminary_results_finetune}
\end{table}

To improve the prediction performance in the downstream task, the downstream users often fine-tune the task header or the graph prompt in a clean dataset on their own. However, this will significantly reduce the attack effectiveness as fine-tuning on clean data may overwrite or mitigate the effects of the backdoor~\cite{sha2022fine}, which is also verified in our experimental results in Tab.~\ref{tab:preliminary_results_finetune}. \method{}/R is the variant of \method{} that without the proposed finetuning-resistant backdoor loss in Eq.~(\ref{eq:finetuning_resistant_backdooring}). From the table, we observe that even simple fine-tuning severely degrades the attack performance, implying the importance of mitigating this effect.

To ensure the effectiveness of backdoor attack after fine-tuning for downstream tasks, following~\cite{cohen2019certified,bansal2022certified}, we propose to adopt adversarial training to make the trojan graph prompt robust against perturbations so that the backdoor remains effective even after fine-tuning the graph prompt and task header.
Specifically, fine-tuning the prompt is equivalent to adding perturbation $\delta$ to $p$. 
Thus, to achieve our goal, we aim to ensure a low backdoor loss for any update $\delta$ with $||\delta||_2\leq\epsilon$, applied to the graph prompt $p$. Simultaneously, the graph prompt $p$ should also maintain low loss for prediction tasks on clean data to preserve the model utility. To achieve the above objectives, we replace the objective function of backdoor poisoning in Eq.~(\ref{eq:train_prompt_header_in_poisoned_dataset}) with the following objective function to optimize the graph prompt $p$ as: 
\begin{equation}
\label{eq:finetuning_resistant_backdooring}
\begin{aligned}
    \min_{p,\theta_{\tau}}\mathcal{L}_{r}(\theta,p+\delta^*,\theta_{\tau}) = & \sum_{v_i\in\mathcal{V}_L}l\left(f_{\tau}[h_{\theta}\{a_p(p,v_i)\}],y_i\right)\\
   + \lambda\sum_{v_i\in\mathcal{V}_P} & l\left(f_{\tau}[h_{\theta}\{a_g(a_p(p+\delta^*,v_i), g_i)\}],y_t\right), \\
   =&\mathcal{L}_c(p)+\lambda\cdot\mathcal{L}_b(p+\delta^*)\\
   s.t.\ \ \delta^*&=\underset{||\delta||\leq\epsilon}{\arg\max}\mathcal{L}_b(p+\delta),
\end{aligned}
\end{equation}
where $\lambda$ is the hyper-parameter to balance model utility and attack performance and $\mathcal{L}_b$ is the finetuning-resistant backdoor loss. With Eq.~(\ref{eq:finetuning_resistant_backdooring}), we will obtain a trojan graph prompt $p^*$ that preserves the backdoor after it is updated within the $\epsilon$-ball, making it remains effective even when the trojan graph prompt and the downstream classifier undergoes fine-tuning. 
Finally, by putting Eq.~(\ref{eq:finetuning_resistant_backdooring}) into Eq.~(\ref{eq:bi_level_optimization_wo_finetuning_resistant}), the overall objective function of \method{} is written as:
\begin{equation}
\begin{aligned}
\label{eq:bi_level_optimization_with_finetuning_resistant}
    &\min_{\theta_g}\mathcal{L}_g(\theta_g,p^*+\delta^*,\theta_{\tau}^*) \quad s.t. \\
 &p^*,\theta_{\tau}^*=\underset{p,\theta_{\tau}}{\arg\min}{\mathcal{L}_{r}(\theta,p+\delta^*,\theta_{\tau})}, \quad \delta^*=\underset{||\delta||\leq\epsilon}{\arg\max}\mathcal{L}_b(p+\delta).
\end{aligned}
\end{equation}




\subsection{Optimization Algorithm}
To efficiently solve the optimization problem in Eq.~(\ref{eq:bi_level_optimization_with_finetuning_resistant}), we propose the following alternating optimization schema.

\noindent\textbf{Lower-level Optimization} We first update $\delta$ with $T$ iterations of gradient ascent with the $\epsilon$-ball to approximate $\delta^*$:
\begin{equation}
\label{updatedelta}    \delta^{t+1}=\delta^{t}+\alpha_{\delta}\nabla_{\delta}\mathcal{L}_b(p+\delta^{t}),
\end{equation}
where the step size is set as $\alpha=\epsilon/(T\cdot\max(||\nabla_{\delta}\mathcal{L}_{b}(p+\delta_t)||_2))$ for optimization under the constraint $||\delta||_2\leq\epsilon$. We then incorporate the approximate $\delta^*$ to update $p$ and $\delta_{\tau}$, respectively:
\begin{equation}
\begin{aligned}
\label{inneroptimization}
    p^{t+1} = p^{t}-\alpha_{p}\nabla_{p}\mathcal{L}_{r}(\theta,p^{t}+\delta^*,\theta_{\tau}^{t})\\
    \theta_{\tau}^{t+1} = \theta_{\tau}^{t}-\alpha_{\tau}\nabla_{\tau}\mathcal{L}_{r}(\theta,p^{t}+\delta^*,\theta_{\tau}^{t}),
\end{aligned}
\end{equation}
where $p^{t}$ and $\theta_{\tau}^{t}$ are the graph prompt and model parameters of task header after $t$ iterations. $\alpha_{p}$ and $\alpha_{\tau}$ are the learning rates for training the graph prompt and the task header.


\noindent\textbf{Upper-level Optimization} In the outer iteration, the updated graph prompt $p^*$, task header parameters $\theta_{\tau}$ are used to approximate $\theta_g^*$, respectively. We then apply first-order approximation~\cite{finn2017model} to compute gradients of $\theta_g$ by: 
\begin{equation}
\begin{aligned}
\label{outteroptimization}
\theta_g^{k+1}=\theta_g^{k}-\alpha_g\nabla_{\theta_g}(\mathcal{L}_g(\theta_g^{k},\overline{p}+\overline{\delta},\overline{\theta_{\tau}}))
\end{aligned}
\end{equation}
where $\overline{p}$, $\overline{\delta}$ and $\overline{\theta_{\tau}}$ indicate gradient propagation stopping, $\theta_g^k$ denotes model parameters after $k$ iterations.
The training algorithm of \method{} is given in Algorithm~\ref{alg:Framwork} in Appendix~\ref{appendix:training_algorithm}. Time complexity analysis is in Appendix~\ref{appendix:time_complexity_analysis}.

\subsection{Training Algorithm}
\label{appendix:training_algorithm}
The {\method} algorithm is detailed in Algorithm \ref{alg:Framwork}. Initially, we randomly initialize the graph prompt $p$, the parameters $\theta_{\tau}$ for the task header $f_{\tau}$, and the parameters $\theta_g$ for the trigger generator $f_g$ (line 1). Next, we train the clean graph prompt $p$ and the task header $f_{\tau}$ by performing gradient descent on $\nabla_p l$ and $\nabla_{\tau} l$, respectively, based on Eq. (\ref{updatedelta}) (line 2).
From lines 3 to 12, we employ a bi-level optimization framework to iteratively optimize the adversarial perturbations $\delta$, graph prompt $p$, task header $f_{\tau}$ and the trigger generator $f_g$ until convergence. Specifically, in the inner loop (lines 4-5), for $t = 1$ to $T$, we update the adversarial perturbation $\delta$ by ascending on the gradient $\nabla_{\delta} \mathcal{L}_b(p + \delta^t)$ according to Equation (\ref{updatedelta}). This step aims to generate perturbations that maximize the backdoor loss $\mathcal{L}_b$.
Subsequently, in the middle loop (lines 7-10), for $t = 1$ to $N$, we update the graph prompt $p$ and the task header parameters $\theta_{\tau}$ by descending on the gradients $\nabla_p \mathcal{L}r$ and $\nabla{\theta_{\tau}} \mathcal{L}_r$, respectively, based on Equation (\ref{inneroptimization}). This step refines the prompt and the task header to minimize the fintuning resistant attack loss $\mathcal{L}_r$, ensuring that the model maintains high performance in a scenario where the downstream header is finetuned.
In the outer loop (line 11), we update the trigger generator parameters $\theta_g$ by descending on the gradient $\nabla_{\theta_g} \mathcal{L}_g$ based on Equation (\ref{outteroptimization}). This step optimizes the trigger generator $f_g$ to produce effective triggers.
Finally, upon convergence of the optimization process, we return the optimized graph prompt $p^* = p$ and the task header $f_{\tau}$ with parameters $\theta_{\tau}$ (line 13).

\begin{algorithm}[t!] 
\caption{Algorithm of {\method}.}
\label{alg:Framwork} 
\begin{algorithmic}[1]
\REQUIRE Graph $\mathcal{G}=(\mathcal{V},\mathcal{E}, \mathbf{X})$, $\lambda$, $\epsilon$.
\ENSURE Trojan graph prompt $p^*$, the task header $f_{\tau}$
\STATE Random initialize graph prompt $p$, $\theta_{\tau}$ for task header $f_{\tau}$, $\theta_{g}$ for trigger generator $f_{g}$
\STATE Train clean graph prompt $p$ and task header $\theta_{\tau}$ by descent on  $\nabla_{p} l$ and $\nabla_{\tau} l$ based on Eq. (\ref{eq:train_clean_prompt_header}).
\WHILE{not converged yet}
    \FOR{t=$1,2,\dots{},T$}
    \STATE Update $\delta$ by ascent on  $\nabla_{\delta} \mathcal{L}_b(p+
    \delta^t)$ based on Eq. (\ref{updatedelta});
    \ENDFOR
    \FOR{t=$1,2,\dots{},N$}
    \STATE Update $p$ by descent on  $\nabla_{p} \mathcal{L}_r$ based on Eq. (\ref{inneroptimization});
    \STATE Update $\theta_{\tau}$ by descent on  $\nabla_{\tau} \mathcal{L}_r$ based on Eq. (\ref{inneroptimization});
    \ENDFOR
    \STATE Update $\theta_g$ by descent on $\nabla_{\theta_g} \mathcal{L}_g$ based on Eq. (\ref{outteroptimization});
\ENDWHILE
    
\RETURN graph prompt $p^*=p$ and task header $f_{\tau}$ with parameter $\theta_{\tau}$;
\label{algorithm}
\end{algorithmic}
\end{algorithm}
\section{Experiments}
In this section, we evaluate the proposed {\method} on various datasets to answer the following research questions: (i) \textbf{RQ1:} How does our \method{} perform in backdoor attacks against various graph prompt learning methods? (ii) \textbf{RQ2:} How robust is our attack performance when the trojan graph prompts and task header are fine-tuned? (iii) \textbf{RQ3:} How does each component of \method{} contribute to the effectiveness in trojan graph prompt attacks?
\subsection{Experimental Setup}
\subsubsection{Datasets}
We conduct experiments on three widely used public real-world datasets, i.e., Cora, Citeseer, and Pubmed~\cite{sen2008collective}, which are standard benchmarks for the semi-supervised node classification downstream task in graph prompt learning~\cite{sun2023graph, zi2024prog}. 
More details of the datasets can be found in Appendix~\ref{appendix:dataset_details}.

\subsubsection{Compared Methods} 
\label{sec:compared_methods}
To the best of our knowledge, \method{} is the first trojan prompt attack against GPL without poisoning GNN encoders. To demonstrate the effectiveness of \method{}, we first introduce a variant as the baseline:
\begin{itemize}[leftmargin=*]
    \item \textbf{BL-Rand}: This is a variant of \method{}. Instead of using adaptive triggers, we inject a fixed subgraph as a universal trigger. The connections of the subgraph are generated based on the Erdos-Renyi (ER) model, and its node features are randomly selected from those in the training graph.
\end{itemize}
Moreover, we adapt several representative state-of-the-art graph backdoor attack methods in supervised setting, including SBA~\cite{zhang2021backdoor}, GTA~\cite{xi2021graph} and UGBA~\cite{dai2023unnoticeable}, to our task. Specifically, we first train the clean graph prompt and task header using Eq.~(\ref{eq:train_clean_prompt_header}). Then, we update the task header on the poisoned dataset. The adaptations are as follows:
\begin{itemize}[leftmargin=*]
    \item \textbf{SBA-P}: It is an adaptation of SBA~\cite{zhang2021backdoor}. In SBA, a static subgraph is injected as a trigger into the training graph for the poisoned node. The subgraph’s connections are based on the Erdos-Renyi (ER) model, and its node features are randomly selected from the training graph. To adapt this to our setting, we fix the prompt graph and the parameters of the pre-trained GNN encoder, updating only the task header during trigger injection.
    \item \textbf{GTA-P}: It is extended from GTA~\cite{xi2021graph}, where a trigger generator crafts subgraphs as triggers tailored to individual samples. The optimization of the trigger generator focuses solely on backdoor attack loss. We adapt GTA similarly to SBA-P to obtain GTA-P.
    \item \textbf{UGBA-P}: It is adapted from UGBA~\cite{dai2023unnoticeable}, which selects representative and diverse nodes as poisoned nodes to fully exploit the attack budget. The adaptive trigger generator in UGBA is optimized with an unnoticeability loss, ensuring that the generated triggers resemble the target nodes. UGBA-P is then constructed following a similar adaptation process.
\end{itemize}

\begin{table*}[t]
    \centering
    \small
    \caption{ Backdoor attack results (CA (\%) | ASR (\%) ). Only clean accuracy is reported for clean graphs.}
    \vskip -1em
    \resizebox{0.72\textwidth}{!}
    {\begin{tabular}{llcccccc}
    \toprule
    {Dataset}  & 
    {Prompts} & Clean Graph  & SBA-P & GTA-P & UGBA-P & BL-Rand & {\method{}}\\
     \midrule
    \multirow{3}{*}{Cora} 
    & {{GraphPrompt} } & 73.7 & 70.4|23.8 & 72.3|28.8 & 72.2|28.9 & 70.8|32.0 & \textbf{73.2|91.3} \\
    & {All-in-one} & 66.2 & 62.3|22.7 & 63.0|24.0 & 63.2|24.0 & 
61.0|25.1 & \textbf{63.7|71.7} \\
    & GPPT & 74.4 & 73.3|39.2 & 73.0|46.5 & 73.4|47.4  & 72.2|53.2 & \textbf{74.4|81.9}\\
    \cmidrule{1-8}
    \multirow{3}{*}{Citeseer} 
    & {{GraphPrompt} } & 64.4 & 63.9|10.3 &  62.2|39.4  & 62.9|13.3  & 63.7|10.7 & \textbf{62.7|94.2}\\
    & {All-in-one} & 62.0 & 61.3|19.0 & 60.8|28.0 & 60.6|28.0 & 59.0|49.0 & \textbf{61.0|75.0} \\
    & GPPT & 62.0 & 59.3|63.6 & 61.7|61.0  & 61.7|60.0  & 59.5|59.7 & \textbf{62.1|73.6}\\
    \cmidrule{1-8}
    \multirow{3}{*}{Pubmed} 
    & {{GraphPrompt} } & 71.0 & 63.8|31.4 &  63.8|21.2 & 70.4|17.3  & 66.5|21.5 & \textbf{70.4|88.3}\\
    & {All-in-one} & 67.6 & 63.6|1.0 & 64.0|14.0 & 64.6|14.0 & 64.2|50.0 & \textbf{64.7|100} \\
    & GPPT & 78.1 & 78.1|44.2 &  78.1|47.2 & 78.1|50.2 & 77.5|51.4 & \textbf{78.1|70.8}\\
    \bottomrule 
    \end{tabular}}
    \vskip -0.8em
    \label{tab:bkd_results_poison_tasker}
\end{table*}

\begin{table*}[t]
    \centering
    \small
    \caption{ Backdoor attack results (CA (\%) | ASR (\%) ) of GraphPrompt in attacking fine-tuned task headers.}
    \vskip -1em
    \resizebox{0.72\textwidth}{!}
    {\begin{tabular}{lccccccc}
    \toprule
    {Dataset} & Clean Graph  & SBA-P & GTA-P & UGBA-P & BL-Rand & \method{}/R & {\method{}}\\
     \midrule
    \multirow{1}{*}{Cora} 
     & 67.6 & 65.6|13.7 &65.7|16.4 &65.6|3.4& 64.1|9.5 & 66.5|41.0 & \textbf{67.5|76.3}\\
    
    \multirow{1}{*}{Citeseer} 
     & 63.9 & 61.7|6.8 & 61.0|24.2 & 60.6|7.9 & 65.9|9.3 & 62.6|45.2& \textbf{63.9|78.1}\\
    \multirow{1}{*}{Pubmed} 
    & 68.7 & 67.1|34.1 & 67.1|26.2 & 66.2|18.6  & 66.7|17.2  & 67.2|37.6 & \textbf{68.2|74.6}\\    
    \bottomrule 
    \end{tabular}}
    \vskip -1.5em
    \label{tab:bkd_results_graphprompt_finetune_tasker}
\end{table*}

\subsubsection{Backbone GNN Models, Pre-training \& GPL Methods}
We conduct experiments targeting GNN encoders within the "pre-training, prompting, and fine-tuning" learning framework. To showcase the flexibility of \method{}, we attack GNNs with different architectures, including GCN \cite{kipf2016semi}, GraphSage \cite{hamilton2017inductive}, and GAT \cite{velivckovic2017graph}. For pre-training, we employ several representative strategies from contrastive and generative learning, such as SimGRACE~\cite{xia2022simgrace}, GraphCL~\cite{you2020graph}, and edge prediction~\cite{hu2020pretraining}. Additionally, we evaluate \method{} against three well-known GPL methods: GraphPrompt~\cite{liu2023graphprompt}, GPPT~\cite{sun2022gppt}, and All-in-one~\cite{sun2023all}, to demonstrate its flexibility across various GPL approaches. Specifically, GraphPrompt and GPPT use token-based prompts, while All-in-one utilizes graph-based prompts. Further details of these GPL methods are given in Appendix~\ref{appendix:more_details_method_GPL}.

\subsubsection{Evaluation Protocol} 
\label{sec:evaluation_protocol}
In this paper, we conduct experiments on downstream supervised node classification under a few-shot setting. In particular, following~\cite{sun2023all}, when applying GraphPrompt and All-in-one, we treat the node classification tasks as graph classification task. Specifically, we construct graph-level datasets by collecting the 2-hop neighbor graphs of nodes from the node-level datasets, and then perform graph classification in the downstream task.
We randomly select 20\% of the nodes from the original dataset as test nodes. Half of these test nodes are used as target nodes for attack performance evaluation, and the other half are used as clean test nodes to evaluate the prediction accuracy of backdoored models on clean samples. From the remaining $80\%$ of nodes, $10\%$ are randomly selected as the validation set. To ensure the prediction performance of GNN encoders on clean graphs, we set the shot number to $100$ by randomly selecting $100$ nodes per class from the remaining nodes.
The average success rate (ASR) on the target node set and clean accuracy (CA) on clean test nodes are used to evaluate the backdoor attacks. Each experiment is conducted 5 times and the average results are reported.

\subsubsection{Implementation Details}
All the details of the implementation of graph prompt learning are provided in Appendix~\ref{appendix:more_GPL_implementation_details}. A 2-layer MLP is deployed as the trigger generator. 
Following the previous efforts~\cite{liu2023graphprompt,sun2022gppt} and ensuring fair comparisons, we set a 2-layer GNN with a sum pooling layer as the backbone encoder for all experiments. For a fair comparison, hyperparameters of all methods are tuned based on the performance of the validation set. 
All models are trained on a Nvidia A6000 GPU with 48GB of memory. 
More details can be found in Appendix~\ref{appendix:implementation_details}.
\subsection{Attack Results}
\subsubsection{Comparisons with Baseline Backdoor Attacks }
\label{maintableexperiment}
In this section, we compare \method{} with baselines on three real-world graphs. We evaluate all three GPL methods: GraphPrompt~\cite{liu2023graphprompt}, GPPT~\cite{sun2022gppt}, and All-in-one~\cite{sun2023all}. The backbone used is GCN \cite{kipf2016semi}, which is pretrained by SimGRACE~\cite{xia2022simgrace}. The results are shown in Tab.~\ref{tab:bkd_results_poison_tasker}. Additional results on various GNN backbones and potential countermeasures are in Appendix~\ref{appendix:more_results_various_GNN_backbones} and~\ref{appendix:potential_defense}, respectively.
From Tab.~\ref{tab:bkd_results_poison_tasker}, we observe:
\begin{itemize}[leftmargin=*]
    \item Existing baselines give poor attack performance in ASR among all datasets and GPL methods. It further validates our claim in Sec.~\ref{sec:limitation_existing_GBA} that existing graph backdoor attacks are ineffective against GPL without poisoning GNN encoders, which implies the necessity of developing trojan graph prompt attacks.
    \item \method{} achieves superior ASR than baselines across all datasets and GPL methods, while maintaining high CA on clean graphs. Especially, \method{} and BL-Rand usually outperform other baselines that are extended from existing backdoor attacks in both ASR and CA. This can be attributed to our strategy of training a clean prompt and task header firstly, followed by our learning objective $\mathcal{L}_{p}$ for trigger generator, which aims to achieve a successful backdoor attack while maintaining clean accuracy.
    This further indicates the effectiveness of our trojan graph prompt attacks
    our proposed trojan graph prompt attacks.
    \item Our \method{} always achieves a significantly higher ASR than BL-Rand in all settings. This indicates the superiority of our feature-aware trigger generator in generating powerful triggers.
    
    
\end{itemize}

\subsubsection{Attacking Fine-Tuned Task Header} 
\label{sec:attacking_finetuned_downstream_tasker}
To answer \textbf{RQ2}, we conduct experiments to explore the attack performance of \method{} where the task header and the trojan graph prompt are fine-tuned, respectively. Specifically, we first fix the trojan graph prompt and fine-tune the task header in the clean graph from the same dataset. More implementation details of fine-tuning task headers are in Appendix~\ref{appendix:implementation_details_finetuning_downstream_tasker}.
GraphPrompt is set as the GPL method, GCN is the GNN backbone.
The results on three datasets are reported in Tab.~\ref{tab:bkd_results_graphprompt_finetune_tasker}. 
From the table, we observe that our \method{} consistently achieves a significantly higher ASR compared to baseline attack methods. Compared to the ASR without fine-tuning task header in Tab.~\ref{tab:bkd_results_poison_tasker}, the ASR drops only slightly, by less than 5\%, while the ASR of \method{}/R that without $\mathcal{L}_r$ significantly drop to $41\%$ on Cora dataset. This demonstrates the effectiveness of our finetuning-resistant loss $\mathcal{L}_r$, which aims to maintain attack performance even if the task header is fine-tuned on a clean dataset. 
Furthermore, our ACC is always comparable to clean accuracy on the clean graph. 
More analysis about the finetuning-resistant loss are in Sec.~\ref{sec:ablation_studies}. Additional results of fine-tuning trojan graph prompt are in Appendix~\ref{appendix:more_results_fientune_graph_prompt}.

\subsubsection{Attacking Frozen Task Header}
To further demonstrate the generalizability of our attack, we evaluate its effectiveness in a more challenging scenario where users train their own task header on a clean dataset, rather than relying on the header provided by the adversary. Specifically, we assume that instead of directly using the attacked task header from the adversary, downstream users train their own task header on the clean graph belonging to the same dataset as the poisoned dataset. The users then apply the trojan graph prompt and the benign task header for the downstream task. We report the results in Tab.~\ref{tab:bkd_results_graphprompt_freeze_tasker}. From the table, we observe that \method{} achieves a significantly higher ASR compared to other baselines while maintaining a clean accuracy comparable to that on a clean graph. Compared with the attack results in Tab.~\ref{tab:bkd_results_poison_tasker}, our ASRs drop slightly. This implies the backdoor embedded in the trojan graph prompts remains effective even when using a benign task header, further demonstrating the generalizability of our attack across different scenarios, highlighting its practical value.

    

\begin{table*}[t]
    \centering
    \small
    \caption{ Backdoor attack results (CA (\%) | ASR (\%) ) of GraphPrompt in attacking freezing task headers. }
    \vskip -1em
    \resizebox{0.66\textwidth}{!}
    {\begin{tabular}{lcccccc}
    \toprule
    {Dataset} & Clean Graph  & SBA-P & GTA-P & UGBA-P & BL-Rand & {\method{}}\\
     \midrule

    \multirow{1}{*}{Cora} & 72.7 & 65.6|13.7 & 65.6|16.4 & 65.6|3.4& 64.1|9.5 & \textbf{72.1|81.0}\\
    \multirow{1}{*}{Citeseer} & 60.1 & 50.3|8.5 & 60.6|2.5  & 59.8|4.2 & 59.6|14.3 & \textbf{59.7|87.5} \\
    \multirow{1}{*}{Pubmed} 
    & 71.0 &     70.5|8.4 & 69.8|17.7 & 71.1|10.0  & 67.1|14.0 & \textbf{70.1|81.0}\\
    
    \bottomrule 
    \end{tabular}}
    \label{tab:bkd_results_graphprompt_freeze_tasker}
\end{table*}


\subsubsection{Attack Performance in Cross-dataset Scenario}
GPL is usually for facilitating rapid adaptation of models to new tasks in diverse application contexts. To demonstrate the feasibility of \method{} in such cross-context scenarios, following~\cite{lyu2024crossba}, we evaluate \method{} in a cross-dataset setting, where the datasets used for pretraining and downstream tasks differ. Specifically, we use Pubmed as the pretraining dataset to train the GNN encoder and Cora and Citeseer for the downstream tasks. To ensure consistency in feature dimensions across datasets, we apply SVD~\cite{abdi2007singular} to reduce the initial feature dimensionality to 100.
The attack results are reported in Tab.~\ref{tab:bkd_results_cross_dataset}.
We observe that even in cross-dataset scenarios, our \method{} achieves ASRs exceeding 80\% in all cases, reaching up to 97.1\%, while maintaining high accuracy on the clean data in downstream tasks. This demonstrates the transferability of \method{} across datasets without compromising model utility in downstream tasks.



\begin{table*}[t]
    \centering
    \small
    \caption{ Backdoor attack results (CA (\%) | ASR (\%) ) in cross-dataset scenarios. Pubmed is used as the pretraining dataset}
    \vskip -1em
    \resizebox{0.78\textwidth}{!}
    {\begin{tabular}{llcccccc}
    \toprule
    {Dataset}  & 
    {Prompts} & Clean Graph  & SBA-P & GTA-P & UGBA-P & BL-Rand & {\method{}}\\
     \midrule
    \multirow{2}{*}{Cora} 
    & {{GraphPrompt} }  & 64.5 & 63.0|12.1 & 63.4|52.8 & 64.5|9.4 & 64.9|14.7 & \textbf{64.5|85.7}\\
    
    & GPPT  & 58.2 & 57.2|59.6 & 58.1|47.6 & 57.8|48.3 & 57.3|47.1 & \textbf{58.1|81.9} \\
    \cmidrule{1-8}
    \multirow{2}{*}{Citeseer} 
    & {{GraphPrompt} }  & 64.4 & 64.0|7.2 & 65.5|56.5 & 64.7|27.9 & 63.3|26.1 & \textbf{65.5|97.1}\\
    & GPPT & 44.9 & 43.1|66.7 & 43.9|36.3 & 44.2|56.4 & 43.8|69.7 & \textbf{44.9|82.3} \\
    \bottomrule 
    \end{tabular}}
    \label{tab:bkd_results_cross_dataset}
\end{table*}


\subsection{Impact of the Size of Trigger Nodes }
In this section, we conduct experiments to explore the attack performance of \method{} given different budgets in the size of trigger graph. Specifically, we vary the size of trigger as $\{1,5,10,15,50\}$. The other settings are the same as the evaluation protocol in Sec. \ref{sec:evaluation_protocol}.
Hyperparameters are selected with the same process as described in Appendix \ref{appendix:implementation_details}.
Fig. \ref{fig:triggersize} shows the results on Cora and Pubmed dataset.
From Fig. \ref{fig:triggersize}, we observe that as the size of the trigger increases, ASR increases in both datasets. While the clean accuracy fluctuates on the Pubmed dataset and slightly decreases on the Cora dataset, the drop remains within 5\%. This indicates that with an increasing budget, our \method{} can continuously achieve higher attack performance with an acceptable drop in clean accuracy.
\begin{figure}
    \centering
    \includegraphics[width=0.85\linewidth]{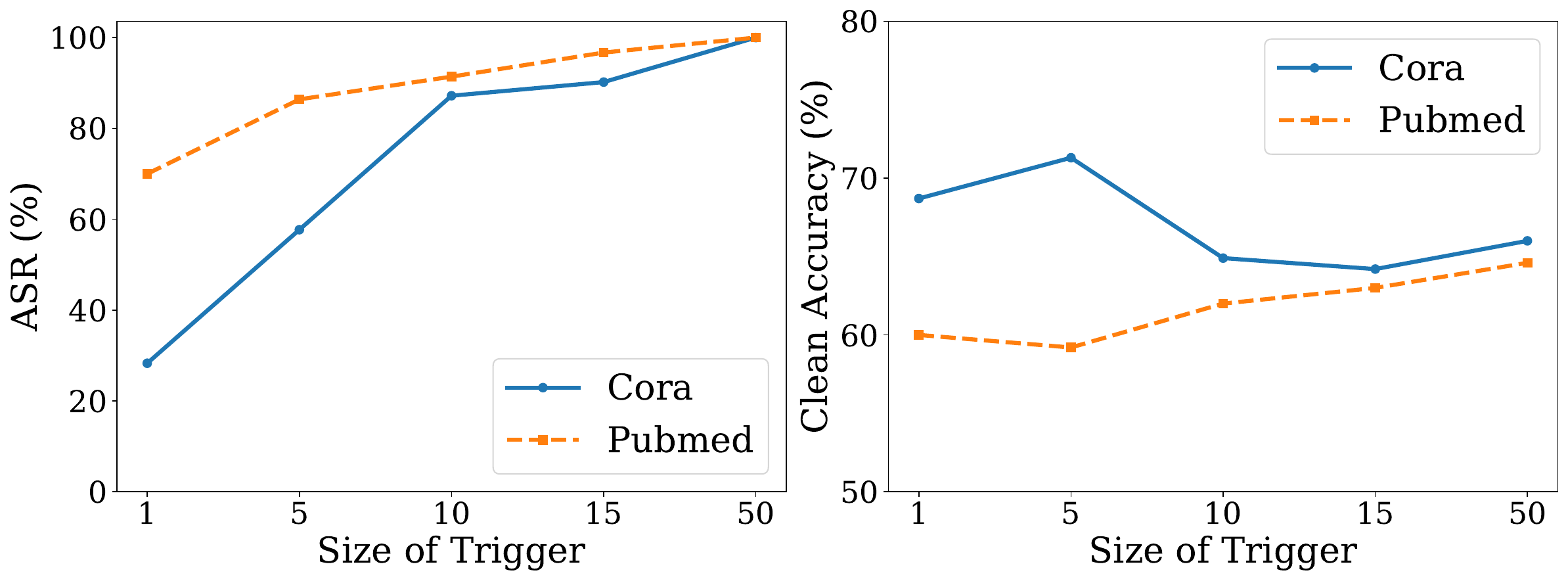}
    \vskip -1em
    \caption{Impact of the Size of Trigger Nodes.}
    \label{fig:triggersize}
\end{figure}
\subsection{Impact of the Ratio of Poisoned Nodes }
We also conduct experiments to assess the attack performance of \method{} under different budgets for the ratio of poisoned nodes. Specifically, we vary the ratio as $\{5\%,10\%,20\%,40\%,100\%\}$. All other settings follow the evaluation protocol described in Sec. \ref{sec:evaluation_protocol}. Hyperparameters are selected according to the procedure outlined in Appendix \ref{appendix:implementation_details}. Fig. \ref{fig:poisonratio} presents the results on the Cora and Pubmed datasets. From Fig. \ref{fig:poisonratio}, we observe that as the poison ratio increases, the attack success rate also increases, while the clean accuracy only changes within a small range. This indicates the superiority of our \method{} in scenarios where the attack budget is sufficient.
\begin{figure}
    \centering
    \includegraphics[width=0.85\linewidth]{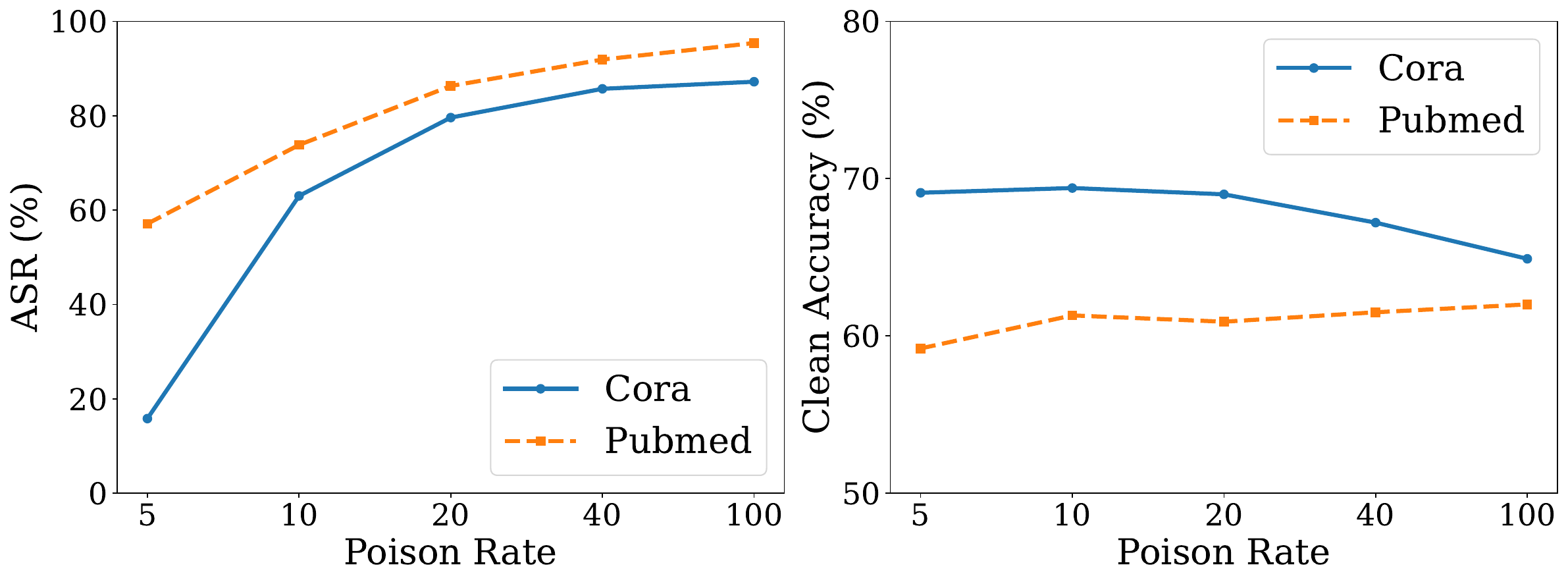}
    \vskip -1em
    \caption{Impact of the Ratio of Poisoned Nodes.}
    \label{fig:poisonratio}
\end{figure}
\subsection{Ablation Studies}
\label{sec:ablation_studies}
\begin{figure}
    \centering
    \vskip -0.5em
    \includegraphics[width=0.85\linewidth]{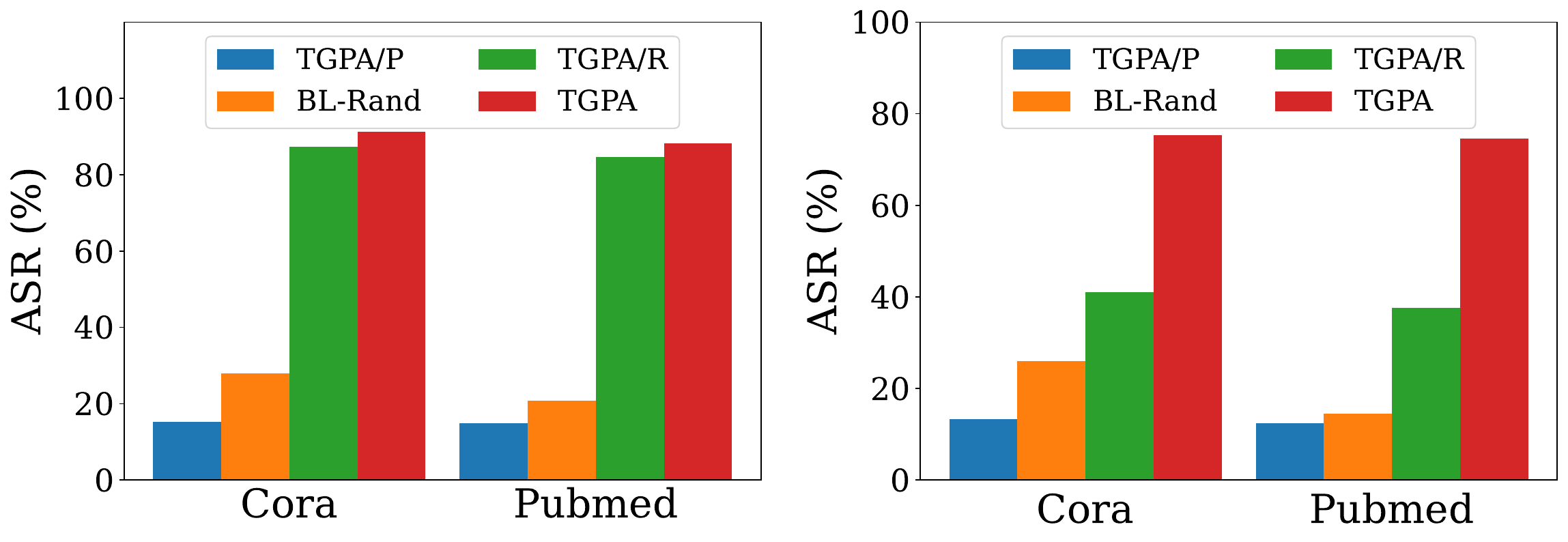}
    \begin{minipage}{0.25\textwidth}
        \centering
        \textbf{(a) W/O finetune}
    \end{minipage}%
    \begin{minipage}{0.15\textwidth}
        \centering
        \textbf{(b) W/ finetune}
    \end{minipage}
    \vskip -1em
    \caption{Ablation study.}
    \label{fig:ablation}
\end{figure}

To answer \textbf{RQ3}, we conduct ablation studies to understand the effects of the proposed trojan graph prompts and the finetuning-resistant loss. The hyperparameter analysis is in Appendix~\ref{appendix:hyperparameter_analysis}.
\textbf{(i)} To demonstrate the effectiveness of the trojan graph prompt poisoning, we obtain a variant \method{/P} by removing the trojan graph prompt optimization in Eq.~(\ref{eq:train_prompt_header_in_poisoned_dataset}) and adding fixed random noises into the clean graph prompt. 
\textbf{(ii)} \method{} deploys a trigger generator to generate a sample-specific trigger for each individual sample. To prove its effectiveness, we consider BL-Rand as a variant, which introduces fixed subgraphs as universal triggers.
\textbf{(iii)} We also involve the variant named \method{/R} in Sec~\ref{sec:attacking_finetuned_downstream_tasker} to explore the effect of finetuning-resistant loss. GraphPrompt is set as the graph prompt method. GCN is the backbone. We conduct experiments both without and with fine-tuning the task header on a clean dataset. For the former, we follow the same experimental setting as in Sec.~\ref{maintableexperiment}; for the latter, we follow the experimental setting in Sec.~\ref{sec:attacking_finetuned_downstream_tasker}. The results on Cora and Pubmed datasets are reported in Fig.~\ref{fig:ablation}. From the figure, we observe that: (i) Both prompt graph optimization and trigger generator optimization can significantly impact ASR, highlighting our idea of optimizing both the prompt and trigger generator effectively aligns with backdoor attacks against graph prompt learning. (ii) In the case where the task header is not fine-tuned on a clean dataset, \method{} still achieves better ASR compared to \method{/R}. This indicates that our fine-tuning resistant loss, $\mathcal{L}_r$, is also effective for attacks when the task header is trained only on the poisoned dataset, enhancing the generalizability of \method{}. (iii) In the scenario where the task header is fine-tuned on a clean dataset, \method{} shows significantly better ASR compared to \method{/R} and other variants. This further demonstrates that the fine-tuning-resistant loss, $\mathcal{L}_r$, helps \method{} achieve robust attack performance in real-world applications.

\section{Conclusion}
In this paper, we investigate the novel problem of trojan graph prompt attacks on GNNs, specifically targeting GPL.
We propose \method{}, a novel backdoor attack framework on GNNs without poisoning GNN encoders. Specifically, to ensure the backdoor effectiveness on trigger-attached samples without affecting model utility, we propose to inject backdoors into graph prompts by carefully crafting trojan graph prompts and triggers. A feature-aware trigger generator is utilized to enhance attack performance.
Additionally, to preserve the effectiveness of backdoors even after downstream fine-tuning, we introduce a finetuning-resistant graph prompt poisoning technique, ensuring that backdoors within trojan graph prompts remain effective despite modifications by end users. Our extensive experiments on various datasets show the high ASR of \method{} on trigger-attached samples while maintaining clean accuracy, validating its effectiveness.
There are two directions that need further investigation. First, in this paper, we only focus on node classification and graph classification on node-level datasets. We will extend the proposed attack to graph-level datasets such as molecule graphs. Second, it is also interesting to investigate how to defend against trojan graph prompt attacks. The discussion of ethical implications is in Appendix~\ref{appendix:ethical_implication}.

\begin{acks}
    This material is based upon work supported by, or in part by the Army Research Office (ARO) under grant number W911NF-21-10198, the Department of Homeland Security (DHS) under grant number 17STCIN00001-05-00, and Cisco Faculty Research Award. The findings in this paper do not necessarily reflect the view of the funding agencies. 
\end{acks}
\newpage
\bibliographystyle{ACM-Reference-Format}
\bibliography{ref}

\newpage
\appendix
\appendix
\newpage
\renewcommand{\theequation}{\thesection.\arabic{equation}}
\setcounter{theorem}{0}
\setcounter{problem}{0}
\setcounter{lemma}{0}
\setcounter{equation}{0}
\setcounter{definition}{0}

\section{More Details of Related Works}
\subsection{Graph Neural Networks} 
Graph Neural Networks (GNNs)~\cite{kipf2016semi,ying2018graph,wang2024efficient, lee2024transitivity,xu2024llm,liang2024survey} have shown great power in modeling graph-structured data, which have been deployed to various applications such as social network analysis~\cite{fan2019graph,dai2021say,liang2024hawkes}, drug discovery~\cite{xu2023graph,dai2021towards,dai2022towards} and energy network analysis~\cite{dai2022graph}. The success of GNNs lies in the message-passing mechanism, which iteratively aggregates a node's neighborhood information to refine the node's representations. For example, GCN~\cite{kipf2016semi} combines a node's neighborhood information by averaging their representations with the target center nodes. To improve the expressivity of GNNs, GIN~\cite{xu2018powerful} further incorporates a hidden layer in combining the neighbors' information. 
Inspired by the recent success~\cite{wei2022chain,lin2024decoding, fang2023hierarchical, fang2023annotations, fang2024not} of prompt learning in natural language processing (NLP) and computer vision (CV), graph prompt learning (GPL) has gained more attention as its promising performance in the transfer learning scenarios. More related works of GPL are introduced in Appendix~\ref{appendix:more_related_works_GPL}.


\subsection{Graph Prompt Learning}
\label{appendix:more_related_works_GPL}
Prompt learning, initially introduced in the language domain, bridges the gap between pre-training and downstream tasks. Recently, graph prompt learning (GPL) has emerged as a promising alternative to traditional fine-tuning methods~\cite{sun2023all, zhao2024all, sun2023graph, li2024zerog, tan2023virtual, yu2024generalized, yu2024hgprompt, liang2024simple}. These techniques leverage shared templates while customizing prompts for specific tasks, enhancing alignment with pre-trained models without modifying the original parameters.

GPL methods can be broadly categorized into two main approaches.
The first approach, \textit{prompt-as-tokens}~\cite{fang2022prompt, fang2024universal, chen2023ultra, gong2023prompt, liu2023graphprompt, ma2024hetgpt, shirkavand2023deep, sun2022gppt}, treats graph prompts as additional features appended to the graph. GraphPrompt~\cite{liu2023graphprompt} introduces a learnable prompt vector in the latent space for downstream tasks. GPPT~\cite{sun2022gppt} divides these prompts into task tokens and structure tokens, reframing node label prediction as predicting a link between a node's structure and task token. 
The second approach, \textit{prompt-as-graphs}~\cite{ge2023enhancing, huang2024prodigy, sun2023all}, introduces prompt tokens as additional nodes with the same representation size as the original nodes~\cite{ge2023enhancing, huang2024prodigy, sun2023all}. For instance, in All-in-One~\cite{sun2023all}, a learnable subgraph of prompt tokens is created, where these tokens have both inner and cross-links to the original graph nodes. This augmented graph is then fed into the pre-trained model, allowing for better integration with the task-specific data.

For downstream tasks, task headers play a crucial role, typically falling into one of two types. Limited-parameter headers employ linear layers with a small number of parameters, while parameter-free headers do not introduce any additional trainable parameters. Instead, they rely on non-parametric operations such as cosine similarity or distance-based metrics to fulfill their roles.

Across different GPL methods, both the graph prompts and the task headers are updated through few-shot learning, making these techniques adaptable to a wide range of downstream tasks with minimal additional data.

\subsection{Backdoor Attacks on Graph}
\label{appendix:more_related_works_backdoor}
In this paper, we focus on graph backdoor attacks that involve attaching backdoor triggers and the target class label to target nodes/graphs \cite{zhang2021backdoor, xi2021graph, dai2023unnoticeable, zhang2024dpgba, zhang2023graph,lyu2024crossba}. A GNN trained on the poisoned graph will then be misled to predict test nodes/graphs attached with trigger to the target class. 
In early efforts, SBA \cite{zhang2021backdoor} introduced a way to inject universal triggers into training data through a subgraph-based method. However, the attack success rate of this approach was relatively low. Building on this, GTA \cite{xi2021graph} presented a method that generated adaptive triggers, adjusting perturbations to each sample individually to improve the effectiveness of the attack. Later, UGBA \cite{dai2023unnoticeable} propose an algorithm for selecting poisoned nodes to make the best use of the attack budget and included an adaptive trigger generator to produce triggers with high cosine similarity to the target node. Recently, DPGBA \cite{zhang2024dpgba} highlighted the limitations of current graph backdoor attacks, citing low success rates or outlier problems. They addressed these by proposing an adversarial learning strategy to generate in-distribution triggers and a new loss function to enhance attack success. GCBA \cite{zhang2023graph} seeks to backdoor graph contrastive learning by attaching pre-defined triggers, thereby creating a backdoored graph. CrossBA~\cite{lyu2024crossba} is a cross-context backdoor attack against GPL, which optimizes the trigger graph and manipulates the pretrained GNN encoder. Krait~\cite{song2024krait} is a concurrent work that injects triggers into GPL to misclassify trigger-attached nodes.

Despite their strong attack performance, these poisoning-based backdoor attacks require training the backdoored model on a poisoned dataset with trigger-attached samples, which is impractical in trojan graph prompt attacks without poisoning GNN encoder. Krait can attack GPL without poisoning GNN encoders, however, it mainly targets~\cite{sun2023all}, which uses subgraphs as prompts, and is not feasible for methods that use learnable tokens as prompts. Krait also fails to address scenarios where graph prompts and task headers are fine-tuned, which affect attack effectiveness.
Our proposed method is inherently different from these method as \textbf{(i)} we propose a trojan graph prompt attack to conduct backdoor attack against GPL without poisoning GNN encoders; \textbf{(ii)} we design a novel finetuning-resistant graph prompt poisoning such that our trojan graph prompt attack remains effective after fine-tuning the graph prompts and task headers.


\section{Time Complexity Analysis}
\label{appendix:time_complexity_analysis}
To analyze the time complexity of the optimization process described in the provided equations, we consider each component of the algorithm separately, focusing on the computational cost associated with updating the adversarial perturbation $\delta$, the graph prompt $p$, the task header parameters $\theta_{\tau}$, and the trigger generator parameters $\theta_{g}$.
Let us define the following variables: 
$n$ is the number of nodes in the graph; 
$m$ is the number of edges; 
$h$ is the embedding dimension; 
$d$ is the average degree of nodes; 
$T$ is the number of iterations for updating $\delta$; 
$N$ is the number of iterations for updating $p$ and $\theta_{\tau}$; 
$K$ is the number of outer iterations for updating $\theta_{g}$; and 
First, we consider the updating of the adversarial perturbation $\delta$. The perturbation $\delta$ is updated over $T$ iterations using gradient ascent as shown in Equation~(\ref{updatedelta}). 
Computing the gradient $\nabla_{\delta} \mathcal{L}_b$ involves a forward and backward pass through the model. Assuming that $\mathcal{L}_b$ is computed over all nodes, the computational cost per iteration is $O(h d n)$. Therefore, the total cost for updating $\delta$ over $T$ iterations is: $O(T \cdot h d n)$.
Next, we analyze the updating of the graph prompt $p$ and task header parameters $\theta_{\tau}$. After obtaining the approximate $\delta^*$, we update $p$ and $\theta_{\tau}$ over $N$ iterations as shown in Equation~(\ref{inneroptimization}).
Computing $\nabla_{p} \mathcal{L}_r$ and $\nabla_{\theta_{\tau}} \mathcal{L}_r$ involves forward and backward passes through the model. The computational cost per iteration is $O(h d n)$, leading to a total cost for updating $p$ and $\theta_{\tau}$ over $N$ iterations of $O(N \cdot h d n)$.

In the outer loop, we update the trigger generator parameters $\theta_{g}$ over $K$ iterations using the first-order approximation, as shown in Equation~(\ref{outteroptimization}):
Computing $\nabla_{\theta_g} \mathcal{L}_g$ involves forward and backward passes through the trigger generator and possibly the model. The computational cost per iteration is $O(h d n)$, resulting in a total cost for updating $\theta_g$ of $O(K \cdot h d n).$
Combining the costs from the lower-level and upper-level optimizations, the total computational cost is $ K \times \left[ O(T \cdot h d n) + O(N \cdot h d n) + O(h d n) \right] = O\left( K (T + N + 1) h d n \right).$
In practice, the constants $T$, $N$, and $K$ are typically small and set by the practitioner based on convergence criteria. 
The overall time complexity of the optimization process is linear with respect to the number of nodes $n$, the average node degree $d$, the embedding dimension $h$, and the total number of iterations $(T + N + K)$. 

\section{Dataset Details}
\label{appendix:dataset_details}
Cora, Citeseer and Pubmed are citation networks where nodes denote papers, and edges depict citation relationships. In these datasets, each node is described using a binary word vector,
indicating the presence or absence of a corresponding word from
a predefined dictionary. In contrast, PubMed employs a TF/IDF
weighted word vector for each node. For all three datasets, nodes
are categorized based on their respective research areas. The dataset statistics are in Tab.~\ref{tab:dataset}.
\begin{table}[t]
    \centering
    \caption{Dataset Statistics}
    \small
    \vskip -1em 
    \begin{tabularx}{0.85\linewidth}{p{0.1\linewidth}CCCc}
    \toprule 
    Datasets & \#Nodes & \#Edges & \#Feature & \#Classes \\
    \midrule
    
    Cora & 2,708 & 5,429 & 1,443 & 7 \\ 
    Citeseer & 3,312 & 9,104 & 3,703 & 6 \\
    Pubmed & 19,717 & 44,338 & 500 & 3 \\ 
    \bottomrule
    \end{tabularx}
    \vskip -2em
    \label{tab:dataset}
\end{table}


\section{More Details of Graph Prompt Learning}
\label{appendix:more_details_method_GPL}

For methods categorized as \textit{prompt as tokens}, given a graph feature matrix $\mathbf{X} = \{x_1, \cdots, x_N\}$ 
$\in \mathbb{R}^{N \times d}$, where $x_i \in \mathbb{R}^{1 \times d}$ is the feature of the $i$-th node and $d$ is the dimension of the feature space. Graph prompt is a learnable vector $p \in \mathbb{R}^{1 \times d}$, which can be added to all node features, making the manipulated feature matrix $\mathbf{X}^* = \{x_1 + p, \cdots, x_N + p\}$. In this way, the reformulated features can replace the original features and process the graph with pre-trained graph models.

\textbf{GraphPrompt} \cite{liu2023graphprompt}  reformulates link prediction to graph pair similarity task. They design prompt token following description above,  while this method assumes the prompt is in the hidden
layer of the graph model. Usually, the hidden size will be
smaller than the original feature, making the prompt token
shorter than the previous one. Furthermore, the graph prompt here is used to assist the graph pooling
operation (also known as Readout). For example, given the node set
$\mathcal{V} = \{v_1, \cdots, v_{|\mathcal{V}|}\}$, the embedding of node $v$ is $\mathbf{h}_v \in \mathbb{R}^{1 \times d}$,
a prompt token $\mathbf{p}_t \in \mathbb{R}^{1 \times d}$ specified to task $t$ is inserted
into the graph nodes by element-wise multiplication ($\otimes$):
$$
\mathbf{s}_t = \text{Readout}(\{\mathbf{p}_t \otimes \mathbf{h}_v : v \in \mathcal{V}\}).
$$

\textbf{GPPT} \cite{sun2022gppt} defines graph prompts as additional tokens, consisting of task tokens and structure tokens. The task token represents the description of the downstream node label to be classified, while the structure token captures the representation of the subgraph surrounding the target node. This setup reformulates the task of predicting node $v$'s label into predicting a potential link between node $v$'s structure token and the task token corresponding to the label.

For methods categorized as \textit{prompt as graphs}, the prompt tokens are some additional nodes that have the same size of node representation as the original nodes.

In \textbf{All-in-one} \cite{sun2023all}: the graph prompt is introduced as a learnable subgraph, efficiently tuned during training. The prompt tokens are additional nodes with the same representation size as the original nodes and are assumed to reside in the same semantic space as the original node features. This allows easy manipulation of node features using these tokens. The token structure consists of two parts: inner links between different tokens and cross-links between the prompt graph and the original graph. These links are pre-calculated using dot products—either between two tokens (inner links) or between a token and an original node (cross-links). The prompt graph is then combined with the original graph through these cross-links, and the resulting graph is fed into the pre-trained model to obtain a graph-level representation.


\section{More Implementation Details}
\subsection{Implementation Details of GPL}
\label{appendix:more_GPL_implementation_details}
We apply SimGRACE~\cite{xia2022simgrace} as the self-supervised learning method to pretrain GNN encoders. We use a learning rate of $0.001$ with a weight decay of $0.0001$ and a 2-layer structure for all GNN backbones. Consistent with the original paper's settings, we use the Adam optimizer. Following the experimental settings in \cite{sun2023all}, during pretraining, we leverage clustering methods to divide the entire graph into 100 distinct subgraphs for creating the pretraining dataset. For downstream tasks, for the GPL methods, GraphPrompt~\cite{liu2023graphprompt} and All-in-one~\cite{sun2023all}, following the settings in~\cite{sun2023all,lyu2024crossba}, we collect the $2$-hop neighborhood graphs as the graph-level datasets, and then convert the node classification tasks to the graph classification tasks.
To train the graph prompts and task header under few-shot learning setting for downstream tasks, the shot number is set to 100.

\subsection{Implementation Details of Backdoor Attacks}
\label{appendix:implementation_details}
A 2-layer MLP is deployed as the trigger generator. 
Following the previous efforts~\cite{liu2023graphprompt,sun2022gppt} and ensuring fair comparisons, we set a 2-layer GNN with a sum pooling layer as the backbone encoder for all experiments. All the hidden dimension is set as $128$. The inner iterations step $N$ is set as $5$ for all the experiments. The trigger size is set as $7$.
All hyperparameters of the compared methods are tuned based. All models are trained on a Nvidia A6000 GPU with 48GB of memory.

\section{Additional results on various GNN backbones}
\label{appendix:more_results_various_GNN_backbones}
In this section, to demonstrate the generalizability of our \method{}, we report the attack results on various GNN backbones, i.e., GraphSage \cite{hamilton2017inductive} and GAT \cite{velivckovic2017graph}.
We follow the experimental setting in Sec.~\ref{maintableexperiment}. Experiments results on Cora and Pubmed datasets are shown in Table~\ref{tab:bkd_results_various_GNN}. We observe that our \method{} consistently achieves a high ASR compared to the baselines while maintaining clean accuracy. This demonstrates the flexibility of our framework.

\begin{table*}[t]
    \centering
    \small
    \caption{ Backdoor attack results (Clean Accuracy (\%) | ASR (\%) ) for various GNN backbones. }
    \vskip -0.8em
    \resizebox{0.76\textwidth}{!}
    {\begin{tabular}{llcccccc}
    \toprule
    {Dataset}  & 
    {Backbone} & Clean Graph  & SBA-P & GTA-P & UGBA-P & BL-Rand & {\method{}}\\
     \midrule
    \multirow{2}{*}{Cora} 
    & GAT  & 68.8 & 67.2|10.1 & 69.0|12.1 & 69.5|4.8 & 66.9|9.4 & 67.4|78.2 \\
    & GraphSage  & 68.4&54.4|6.5 &65.6|13.2 & 55.2|5.7&66.7|33.7 & 68.3|72.1\\
    \cmidrule{1-8}
    \multirow{2}{*}{Pubmed} 
    & GAT  & 67.6 & 68.0|24.5 & 68.0|43.8 & 67.2|10.7 & 68.6|25.1 & 68.1|87.7 \\
    & GraphSage  &  60.4& 56.6|19.1 & 57.5|13.2 & 59.7|34.7& 60.8|32.4& 59.6|74.5\\
    \bottomrule 
    \end{tabular}}
    \label{tab:bkd_results_various_GNN}
\end{table*}

\section{Additional results in the fine-tuning scenario}
\subsection{Implementation Details of Fine-tuning Task Headers}
\label{appendix:implementation_details_finetuning_downstream_tasker}
\textbf{GraphPrompt}~\cite{liu2023graphprompt} applies a parameter-free task header. Specifically, for each class in downstream tasks, it will accumulate all representations of nodes in the same class to obtain a representative center $h_{\theta}(c_i)$ for each class $i$. In the inference phase, given a representation $h_{\theta}(v)$ of target node $v$, it will be predicted as class $i$ if the similarity between $h_{\theta}(v)$ and $h_{\theta}(c_i)$ is largest. Therefore, to fine-tune the task headers of GraphPrompt, we add the representations of the new added nodes on the clean dataset and update the representative center.

For the task headers of other GPL methods that have limited parameters~\cite{sun2023all,sun2022gppt}, we can simply update their parameters during fine-tuning.
\subsection{Attack Performance on Fine-tuned Trojan Graph Prompts}
\label{appendix:more_results_fientune_graph_prompt}
In this subsection, we will explore the performance of \method{} in the scenario of fine-tuning trojan graph prompts. Specifically, we fix the task header and fine-tune trojan graph prompt on the clean graph from the same dataset as the poisoned graph. We also explore the scenario that both the trojan graph prompt and the task header are fine-tuned. The attack results on Cora dataset are reported in Tab.~\ref{tab:finetune_trojan_graph_prompt}. 
From the table, we observe that our \method{} can maintain both high ASR on trigger-attached nodes and high accuracy on clean nodes in all settings that fine-tuning both trojan graph prompt and task header. This demonstrates the robustness of our trojan graph prompt against fine-tuning, further implying the effectiveness of our finetuning-resistant backdoor loss in Eq.~(\ref{eq:finetuning_resistant_backdooring}).

\begin{table}[t]
    \centering
    \caption{Attack results of on Cora before and after fine-tuning trojan graph prompts (ASR (\%)|CA (\%)).}
    \small
    \vskip -1em 
    \begin{tabularx}{0.9\linewidth}{lcCc}
    \toprule 
     & Clean Graph & \method{}/R & \method{} \\
    \midrule
    
    W/o finetune & 73.7 & 73.2|88.6 & \textbf{73.2|91.3}\\
    Prompt &  70.8 & 69.9|85.4 & \textbf{70.7|89.7}\\
    Prompt
    \& Header &  67.4 & 66.6|39.1 & \textbf{67.3|80.1}\\
    \bottomrule
    \end{tabularx}
    \vskip -1em
    \label{tab:finetune_trojan_graph_prompt}
\end{table}


\subsection{Hyperparameter Analysis}
\label{appendix:hyperparameter_analysis}
\begin{figure}
    \centering
    \includegraphics[width=0.85\linewidth]{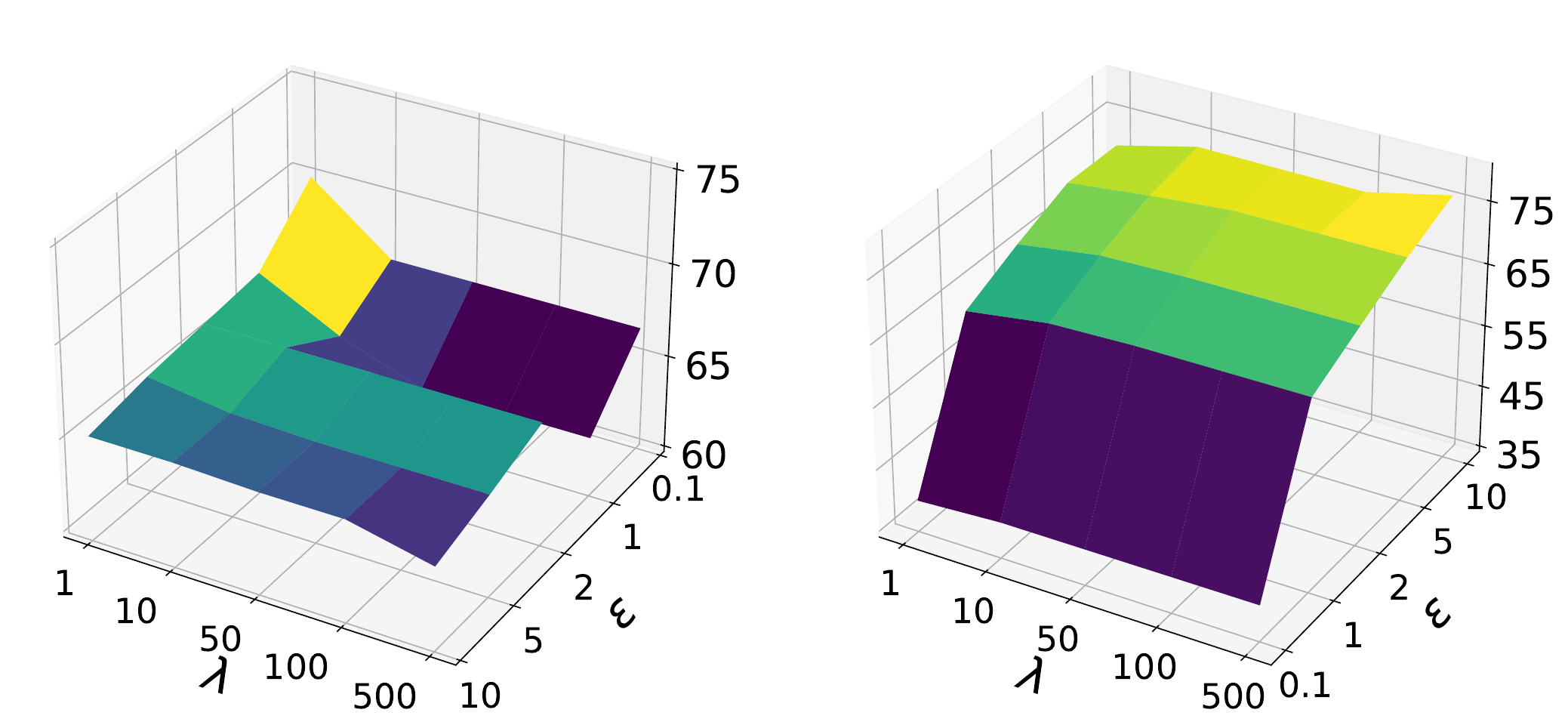}
    \begin{minipage}{0.2\textwidth}
        \centering
        \textbf{(a) CA}
    \end{minipage}%
    \begin{minipage}{0.2\textwidth}
        \centering
        \textbf{(b) ASR}
    \end{minipage}
    \vskip -1em
    \caption{Hyperparameter analysis.}
    \label{fig:hyperana}
\end{figure}

In this subsection, we further investigate how the hyperparameter $\lambda$ and $\epsilon$ affect the performance of \method{}, where $\lambda$ controls the contributions of the proposed finetuning-resistant backdoor loss in Eq.~(\ref{eq:finetuning_resistant_backdooring}) and $\epsilon$ controls the magnitude of changes to the parameters of $p$ in finetuning-resist graph prompt poisoning. More specifically, we vary $\lambda$ and $\epsilon$ as $\{1, 10, 50, 100, 500\}$ and $\{0.1, 1, 2, 5, 10\}$, respectively. We report the ASR and CA on the Cora dataset. The other experimental setting are the same as that in Sec.~\ref{sec:evaluation_protocol}. GraphPrompt is the GPL method. The attack results of \method{} after fine-tuning the task header are shown in Fig.~\ref{fig:hyperana}. From this figure, we find that ASR will increase with the increasing of $\lambda$ and $\epsilon$. As for the CA, it will first decrease and then stay consistent if $\lambda$ and $\epsilon$ is too high. From the figure, we can observe that when $\lambda=50$ and $\epsilon=5$, we will achieve the best balance between the attack performance and model utility.


\section{Potential Countermeasures}
\label{appendix:potential_defense}
In this section, we investigate potential countermeasures against the Trojan Graph Prompt Attack (TGPA). As TGPA is a novel type of graph backdoor attack, no prior work has specifically examined defenses for this scenario. One preliminary defense strategy involves pruning edges with low cosine similarity scores~\cite{dai2023unnoticeable}. Accordingly, we assess the effectiveness of this approach against \method. Following~\cite{dai2023unnoticeable}, we remove edges whose cosine similarity scores are below 0.1 and 0.4 in the Cora and Pubmed datasets, respectively, during the inference phase. This pruning is performed after training the poisoned graph prompts and task headers. Note that GraphPrompt and GCN serve as the backbone GPL and GNN architectures, respectively.
Table~\ref{tab:potential_defense} presents the experimental results on both datasets. From these results, we observe the following: (i) Despite pruning dissimilar edges, \method{} still achieves a high ASR. This outcome is reasonable, as the attacker has already embedded backdoors into the graph prompts prior to pruning, effectively executing a poisoning attack that misclassifies test samples. (ii) On the Pubmed dataset, TGPA attains $62.4\%$ classification accuracy (CA) on the poisoned graph, compared with $71.4\%$ CA on the clean graph, indicating that pruning significantly reduces model utility. These findings show the need for more advanced and effective defense mechanisms against \method{}.

\begin{table}[t]
    \centering
    \small
    \caption{ Backdoor attack results (Clean Accuracy (\%) | ASR (\%) ) for \method{} with potential defense strategies. }
    \vskip -0.8em
    \resizebox{0.76\linewidth}{!}
    {\begin{tabular}{llcc}
    \toprule
    {Dataset}  & 
    {Defense} & Clean Graph  & {\method{}}\\
     \midrule
    \multirow{2}{*}{Cora} 
    & None  & 73.7 & 73.2|91.3 \\
    & Prune  & 72.8 & 68.6|93.6\\
    \midrule
    \multirow{2}{*}{Pubmed} 
    & None  & 71.0 & 70.4|88.3 \\
    & Prune  & 71.4 & 62.4|91.4\\
    \bottomrule 
    \end{tabular}}
    \label{tab:potential_defense}
\end{table}

\section{Ethical Implications}
\label{appendix:ethical_implication}
In this paper, we study a novel trojan graph prompt attack against graph prompt learning without poisoning GNN encoders. Our work uncovers the vulnerability of GNNs with fixed model parameters to the trojan graph prompt attack. Our work aims to raise awareness about the robustness issue inherent in GNNs and inspire the following works to develop more advanced backdoor defenses to protect GPL against trojan graph prompt attack. All datasets we used in this paper are publicly available, no sensitive or private dataset from individuals or organizations was used. Our work is mainly for research purposes and complies with ethical standards. Therefore, it does not have any negative ethical impact on society.

    

\end{document}